\documentclass[final,5p,times,twocolumn]{elsarticle}

\usepackage{amssymb}
\usepackage{amsmath}

\usepackage{graphicx}
\usepackage{caption}

\usepackage{multirow}
\usepackage{booktabs}
\usepackage{adjustbox}
\usepackage{url}

\journal{*************}

\begin{document}
\captionsetup[figure]{labelfont={bf},name={Fig.},labelsep=period}

\begin{frontmatter}



\title{Modeling Sentiment Dependencies with Graph Convolutional Networks for Aspect-level Sentiment Classification}


\author[1]{Pinlong Zhao}
\ead{pinlongzhao@tju.edu.cn}
\author[2]{Linlin Hou}
\ead{llhou@mail.nankai.edu.cn}

\author[1]{Ou Wu}
\ead{wuou@tju.edu.cn}

\address[1]{Center for Applied Mathematics, Tianjin University}
\address[2]{Center for Combinatorics, Nankai University}

\begin{abstract}
Aspect-level sentiment classification aims to distinguish the sentiment polarities over one or more aspect terms in a sentence. Existing approaches mostly model different aspects in one sentence independently, which ignore the sentiment dependencies between different aspects. However, we find such dependency information between different aspects can bring additional valuable information. In this paper, we propose a novel aspect-level sentiment classification model based on graph convolutional networks (GCN) which can effectively capture the sentiment dependencies between multi-aspects in one sentence.  Our model firstly introduces bidirectional attention mechanism with position encoding to model aspect-specific representations between each aspect and its context words, then employs GCN over the attention mechanism to capture the sentiment dependencies between different aspects in one sentence. We evaluate the proposed approach on the SemEval 2014 datasets. Experiments show that our model outperforms the state-of-the-art methods. We also conduct experiments to evaluate the effectiveness of GCN module, which indicates that the dependencies between different aspects is highly helpful in aspect-level sentiment classification.
\end{abstract}

\begin{keyword}
Sentiment classification \sep Aspect-level \sep Sentiment dependencies \sep Graph convolutional networks


\end{keyword}

\end{frontmatter}


\section{Introduction}

Aspect-level sentiment classification \cite{BoPang2008,BingLiu2012} is a fundamental natural language processing task that gets lots of attention in recent years. It is a fine-grained task in sentiment analysis, which aims to infer the sentiment polarities of aspects in their context. For example, in the sentence \textit{``The price is reasonable although the service is poor"}, the sentiment polarities for the two aspect terms, \textit{``price"} and \textit{``service"}, are positive and negative respectively. An aspect term (or simply aspect) is usually an entity or an entity aspect.

Aspect-level sentiment classification is much more complicated than sentence-level sentiment classification, because identifying the parts of sentence describing the corresponding aspects is difficult. Traditional approaches \cite{SvetlanaKiritchenko2014,wagner2014dcu} mainly focus on statistical methods to design a set of handcrafted features to train a classifier (e.g., Support Vector Machine). However, such kind of feature-based work is labor-intensive. In recent years, neural network models \cite{PoriaS2016,DuyuTang2016} are of growing interest for their capacity to automatically generate useful low dimensional representations from aspects and their contexts, and achieve great accuracy on the aspect-level sentiment classification without careful engineering of features. Especially, by the ability to effectively identify which words in the sentence are more important on a given aspect, attention mechanisms \cite{Volodymyr2014,Dzmitry2015} implemented by neural networks are widely used in aspect-level sentiment classification \cite{YequanWang2016Attention,chenPeng2017,Ma2017interactive,Ma2018,YouweiSong2019,RuixinMa2019Feature}.  Chen et al. \cite{chenPeng2017} model a multiple attention mechanism with a gated recurrent unit network to capture the relevance between each context word and the aspect. Ma et al. \cite{Ma2017interactive} design a model which learns the representations of the aspect and context interactively with two attention mechanisms. Song et al. \cite{YouweiSong2019} propose an attentional encoder network, which employ multi-head attention for the modeling between context and aspect. These attention-based models have proven to be successful and effective in learning aspect-specific representations.

\begin{figure}
	\centering
		\includegraphics[scale=.58]{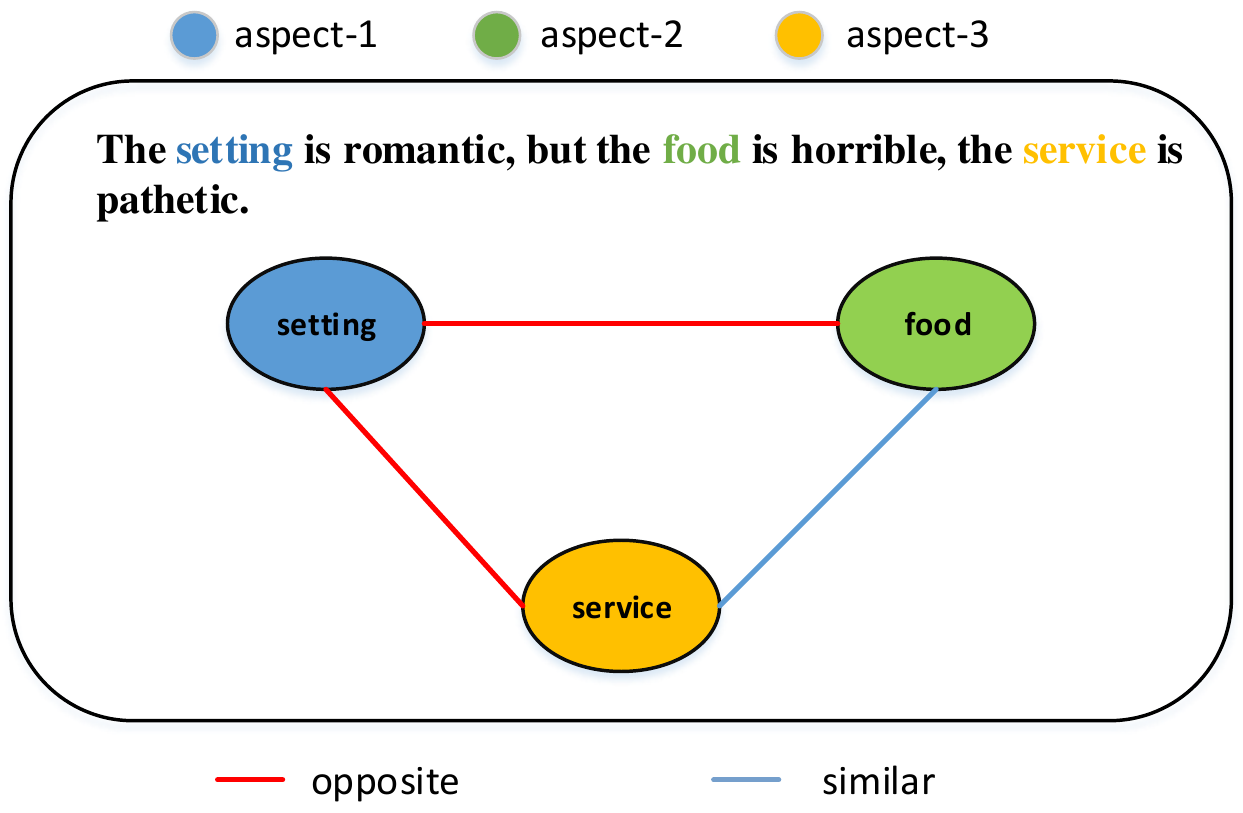}
	\caption{An example to illustrate the usefulness of the sentiment dependencies between multiple aspects. The dependencies can be inferred by some knowledge in the sentence, e.g., conjunction. The evidence of the usefulness of the sentiment dependencies is that we can easily guess the true sentiment of \textit{``food"} even if we mask the word \textit{``horrible"}.}
	\label{FIG:1}
\end{figure}

Despite these advances, the studies above still remain problems. They all build models with each aspect individually ignoring the sentiment dependencies information between multiple aspects, which will lose some additional valuable information. For example, as we can see from the example given in Fig.~\ref{FIG:1}, the sentiment polarity of the first aspect \textit{``setting"} is positive. From the conjunction \textit{``but"}, we are easy to know that the second aspect \textit{``food"} has opposite sentiment polarity with \textit{``setting"}. By this sentiment dependency relation, we can guess the polarity of aspect \textit{``food"} is negative. Similarly, from the second comma, we conjecture that the sentiment polarity of the last aspect \textit{``service"} is likely the same as \textit{``food"}. Therefore, the sentiment dependencies are helpful to infer the sentiment polarities of aspects in one sentence.

In this paper, we propose a novel method to model \textbf{S}entiment \textbf{D}ependencies with \textbf{G}raph \textbf{C}onvolutional \textbf{N}etworks (SDGCN) for aspect-level sentiment classification. GCN is a simple and effective convolutional neural network operating on graphs, which can catch inter-dependent information from rich relational data \cite{Kipf2016}. For every node in graph, GCN encodes relevant information about its neighborhoods as a new feature representation vector. In our case, an aspect is treated as a node, and an edge represents the sentiment dependency relation of two nodes. Our model learns the sentiment dependencies of aspects via this graph structure. As far as we know, our work is the first to consider the sentiment dependencies between aspects in one sentence for aspect-level sentiment classification task. Furthermore, in order to capture the aspect-specific representations, our model applies bidirectional attention mechanism with position encoding before GCN. We evaluate the proposed approach on the SemEval 2014 datasets. Experiments show that our model outperforms the state-of-the-art methods.\footnote{Source code is available at \url{https://github.com/Pinlong-Zhao/SDGCN}.} The main contributions of this paper are presented as follows:
\begin{itemize}
\item To the best of our knowledge, this is the first study to consider the sentiment dependencies between aspects in one sentence for aspect-level sentiment classification.
\item We design bidirectional attention mechanism with position encoding to capture the aspect-specific representations.
\item We propose a novel multi-aspects sentiment classification framework, which employs GCN to effectively capture the sentiment dependencies between different aspects in one sentence.
\item We evaluate our method on the SemEval 2014 datasets. And experiments show that our model achieves superior performance over the state-of-the-art approaches.
\end{itemize}

\section{Related work}
In this section, we will review related works on aspect-level sentiment classification and graph convolutional network briefly.
\subsection{Aspect-level sentiment classification}
Sentiment analysis, also known as opinion mining \cite{Kim2014Convolutional,AbidAYL19Fazeel}, is an important research topic in Natural Language Processing (NLP). Aspect-level sentiment classification is a fine-grained task in sentiment analysis.
In aspect-level sentiment classification , early works mainly focus on extracting a set of features like bag-of-words features and sentiment lexicons features to train a sentiment classifier \cite{rao2009semi}. These methods including rule-based methods \cite{ding2008holistic} and statistic-based methods \cite{Jiang2011Target} rely on feature-engineering which are labor intensive. In recent years, deep neural network methods are getting more and more attention as they can generate the dense vectors of sentences without handcrafted features \cite{Dong2014Adaptive,Hai2015Phrasernn}. And the vectors are low-dimensional word representations with rich semantic information remained. Moreover, using the attention mechanism can enhance the sentence representation for concentrating on the key part of a sentence given an aspect \cite{Li2017PDeep,Li2018Transformation,XiaMa2019Modeling}. Wang et al. \cite{YequanWang2016Attention} propose ATAE-LSTM that combines LSTM and attention mechanism. The model makes embeddings of aspects to participate in computing attention weights. RAM is proposed by Chen et al. \cite{chenPeng2017} which adopts multiple-attention mechanism on the memory built with bidirectional LSTM. Ma et al. \cite{Ma2017interactive} design a model with the bidirectional attention mechanism, which interactively learns the attention weights on context and aspect words respectively. Song et al. \cite{YouweiSong2019} propose an attentional encoder network, which eschews recurrence and apply multi-head attention for the modeling between context and aspect. However, these attention works model each aspect separately in one sentence, which may loss some sentiment dependency information on multiple aspects case.

\subsection{Graph convolutional network}
Graph convolutional network \cite{Bruna2014Transformation} is effective at dealing with graph data which contains rich relation information. Many works dedicate to extending GCN for image tasks \cite{Henaff2015Deep,Defferrard2016Convolutional,Qi20173D,Li2018Factorizable}.  Chen et al. \cite{Chen2019Multi} build the model via GCN for multi-label image recognition, which propagates information between multiple labels and consequently learns inter-dependent classifiers for each of image labels. GCN has also received growing attention in NLP recently such as semantic role labeling \cite{Marcheggiani2017Encoding}, machine translation \cite{Bastings2017Graph} and relation classification \cite{Li2019Classifying}. Some works explore graph neural networks for text classification \cite{Hao2018Large,Yue2018Sentence}. They view a document, a sentence or a word as a graph node and rely on the relation of nodes to construct the graph. The studies above show that GCN can effectively capture relation between nodes. Inspired by these, we adopt GCN to get the sentiment dependencies between multi-aspects.

\section{Methodology}

\begin{figure*}
	\centering
		\includegraphics[scale=.55]{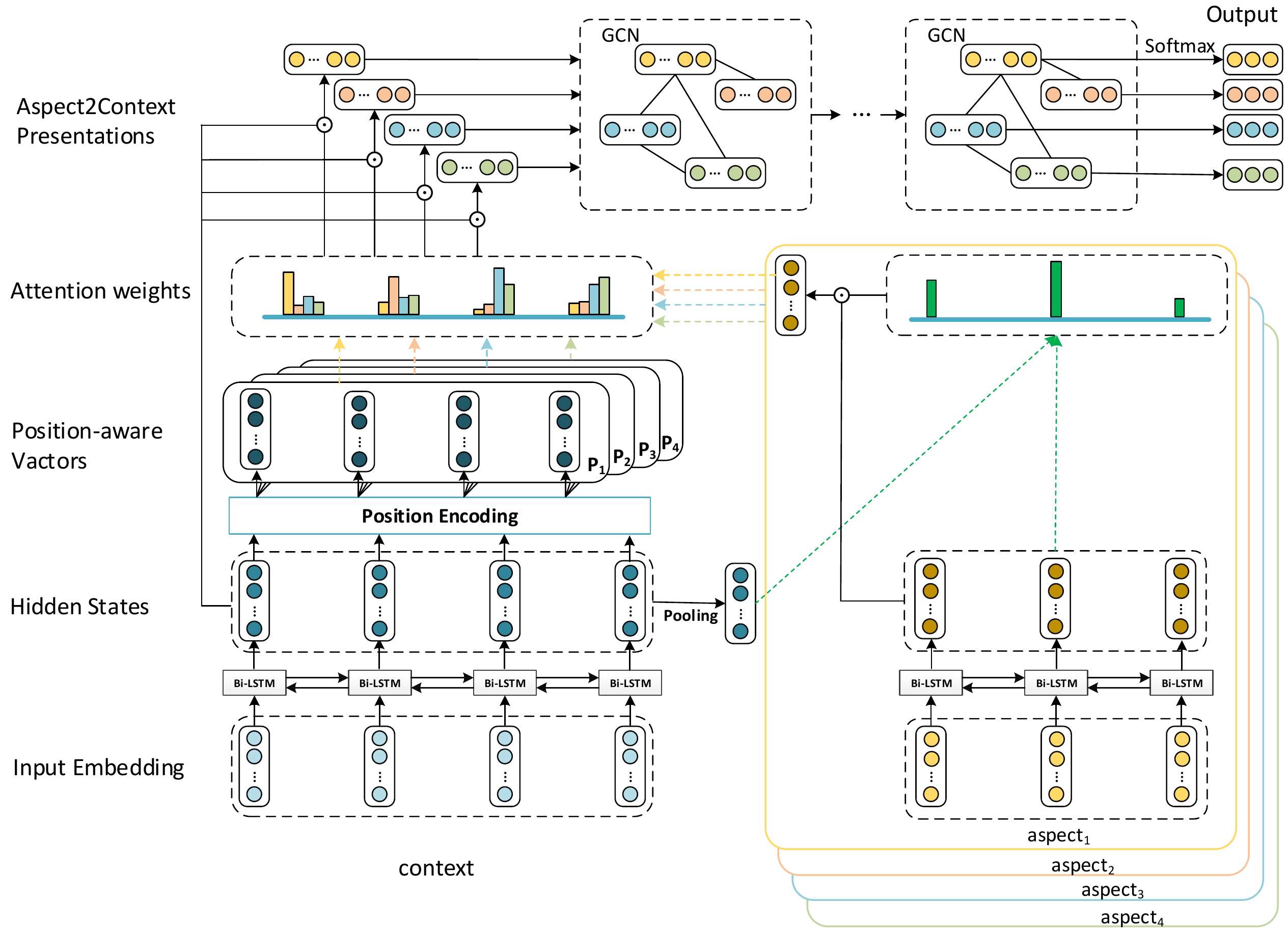}
	\caption{An example to illustrate the usefulness of the sentiment dependencies between multiple aspects. The dependencies can be inferred by some knowledge in the sentence, e.g., conjunction. The evidence of the usefulness of the sentiment dependencies is that we can easily guess the true sentiment of ¡°food¡± even if we mask the word ¡°horrible¡±.}
	\label{FIG:2}
\end{figure*}

Aspect-level sentiment classification can be formulated as follows. Given an input context consists of $N$ words $W^c=\{w^c_1,w^c_2,\ldots,w^c_N\}$, and $K$ aspect teams $W^a=\{W^{a_1},W^{a_2},\ldots,W^{a_K}\}$. Each aspect $W^{a_i}=\{w^{a_i}_1,w^{a_i}_2,\ldots,w^{a_i}_{M_i}\}$ is a subsequence of sentence $W^c$, which contains $M_i\in [1,N)$ words. It is required to construct a sentiment classifier that predicts the sentiment polarities of the multiple aspect teams.

We present the overall architecture of the proposed SDGCN in Fig.~\ref{FIG:2}. It consists of the input embedding layer, the Bi-LSTM, the position encoding, the bidirectional attention mechanism, the GCN and the output layer. Next, we introduce all components sequentially from input to output.

\subsection{Input embedding layer }
Input embedding layer maps each word to a high dimensional vector space. We employ the pretrained embedding matrix GloVe \cite{Pennington2014Glove} and pretrained model BERT \cite{Devlin2019BERT} to obtain the fixed word embedding of each word. Then each word will be represented by an embedding vector $e_t\in \mathbb{R}^{d_{emb}\times1}$ , where $d_{emb}$ is the dimension of word vectors. After embedding layer, the context embedding is denoted as a matrix $E^c\in \mathbb{R}^{d_{emb}\times N}$, and the $i$-th aspect embedding is denoted as a matrix $E^{a_i}\in \mathbb{R}^{d_{emb}\times M_i}$.

\subsection{Bidirectional Long Short-Term Memory (Bi-LSTM)}
We employ Bi-LSTM on top of the embedding layer to capture the contextual information for each word. After feeding word embedding to Bi-LSTM, the forward hidden state $\overrightarrow{h_t}\in \mathbb{R}^{d_{hid}\times 1}$ and the backward hidden state $\overleftarrow{h_t}\in \mathbb{R}^{d_{hid}\times 1}$ are obtained, where $d_{hid}$ is the number of hidden units. We concatenate both the forward and the backward hidden state to form the final representation:
\begin{flalign}\label{1}
&h_t=[\overrightarrow{h_t},\overleftarrow{h_t}]\in \mathbb{R}^{2d_{hid}\times 1}&
\end{flalign}

In our model, we employ two Bi-LSTM separately to get the sentence contextual hidden output $H^c=[h^c_1,h^c_2,\ldots,h^c_N]\in \mathbb{R}^{2d_{hid}\times N}$ and each aspect contextual hidden output $H^a_i=[h^{a_i}_1,h^{a_i}_2,\ldots,h^{a_i}_{M_i}]\in \mathbb{R}^{2d_{hid}\times M_i}$. Note that, the Bi-LSTM for each different aspect shares the parameters.

\subsection{Position encoding}
Based on the intuition that the polarity of a given aspect is easier to be influenced by the context words with closer distance to the aspect, we introduce position encoding to simulate this normal rules in natural language. Formally, given an aspect $W^{a_i}$ that is one of the $K$ aspects, where $i\in[1,K]$ is the index of aspects, the relative distance $d^{a_i}_t$ between the $t$-th word and the $i$-th aspect is defined as follows:
\begin{flalign}\label{2}
&d^{a_i}_t=
\begin{cases}
1, & \text{$dis=0$}\\
1-\frac{dis}{N}, & \text{$1\leq dis\leq s$}\\
0, & \text{$dis> s$}
\end{cases}&
\end{flalign}
where $dis$ is the distance between a context word and the aspect (here we treat an aspect as a single unit, and $d = 0$ means that the context word is also the aspect word), $s$ is a pre-specified constant, and $N$ is the length of the context. Finally, we can obtain the position-aware representation with position information:
\begin{flalign}\label{3}
&p^{a_i}_t=d^{a_i}_th^c_t&\nonumber\\
&P^{a_i}=P_i=[p^{a_i}_1,p^{a_i}_2,\ldots,p^{a_i}_N]&
\end{flalign}

\subsection{Bidirectional attention mechanism}
In order to capture the interactive information between the context and the aspect, we employ a bidirectional attention mechanism in our model. This mechanism consists of two modules: \textit{context to aspect attention} module and \textit{aspect to context attention} module. Firstly, the former module is used to get new representations of aspects based on the context. Secondly, based on the new representations, the later module is employed to obtain the aspect-specific context representations which will be fed into the downstream GCN.
\subsubsection{Context to aspect attention }
\textit{Context to aspect attention} learns to assign attention weights to the aspect words according to a query vector, where the query vector is $\overline{h^c}\in \mathbb{R}^{2d_{hid}\times 1}$  which is obtained by average pooling operation over the context hidden output $H^c$ . For each hidden word vector $h^{a_i}_t\in \mathbb{R}^{2d_{hid}\times 1}$ in one aspect, the attention weight $\beta^{a_i}_t$ is computed as follows:
\begin{flalign}\label{4}
&f_{ca}(\overline{h^c},h^{a_i}_t)=\overline{h^c}^T\cdot W_{ca}\cdot h^{a_i}_t&
\end{flalign}
\begin{flalign}\label{5}
&\beta^{a_i}_t=\frac{exp(f_{ca}(\overline{h^c},h^{a_i}_t))}{\sum^{M_i}_{t=1}exp(f_{ca}(\overline{h^c},h^{a_i}_t))}&
\end{flalign}
where $W_{ca}\in \mathbb{R}^{2d_{hid}\times2d_{hid}}$ is the attention weight matrix.\par
After computing the word attention weights, we can get the weighted combination of the aspect hidden representation as a new aspect representation:
\begin{flalign}\label{6}
&m^{a_i}=\sum^{M_i}_{t=1}\beta^{a_i}_t\cdot h^{a_i}_t&
\end{flalign}
\subsubsection{Aspect to context attention }
\textit{Aspect to context attention} learns to capture the aspect-specific context representation, which is similar to \textit{context to aspect attention}. Specifically, the attention scores is calculated by the new aspect representation $m^{a_i}$ and the position-aware representation $p^{a_i}_t$. The process can be formulated as follows:
\begin{flalign}\label{7}
&f_{ac}(m^{a_i},p^{a_i}_t)={m^{a_i}}^T\cdot W_{ac}\cdot p^{a_i}_t&
\end{flalign}
\begin{flalign}\label{8}
&\gamma^{a_i}_t=\frac{exp(f_{ac}(m^{a_i},p^{a_i}_t))}{\sum^{N}_{t=1}exp(f_{ac}(m^{a_i},p^{a_i}_t))}&
\end{flalign}
\begin{flalign}\label{9}
&x^{a_i}=x_i=\sum^{N}_{t=1}\gamma^{a_i}_t\cdot h^c_t&
\end{flalign}
where $W_{ac}\in \mathbb{R}^{2d_{hid}\times 2d_{hid}}$ is the attention weight matrix. By now, we get the aspect-specific representations $X=[x_1,x_2,\ldots,x_K]$ between each aspect and its context words, where K is the number of aspects in the context.
\subsection{Graph convolutional network}
GCN is widely used to deal with data which contains rich relationships and interdependency between objects, because GCN can effectively capture the dependence of graphs via message passing between the nodes of graphs. We also employ a graph to capture the sentiment dependencies between aspects. The final output of each GCN node is designed to be the classifier of the corresponding aspect in our task. Moreover, there are no explicit edges in our task. Thus, we need to define the edges from scratch.

\begin{figure}
	\centering
		\includegraphics[scale=.85]{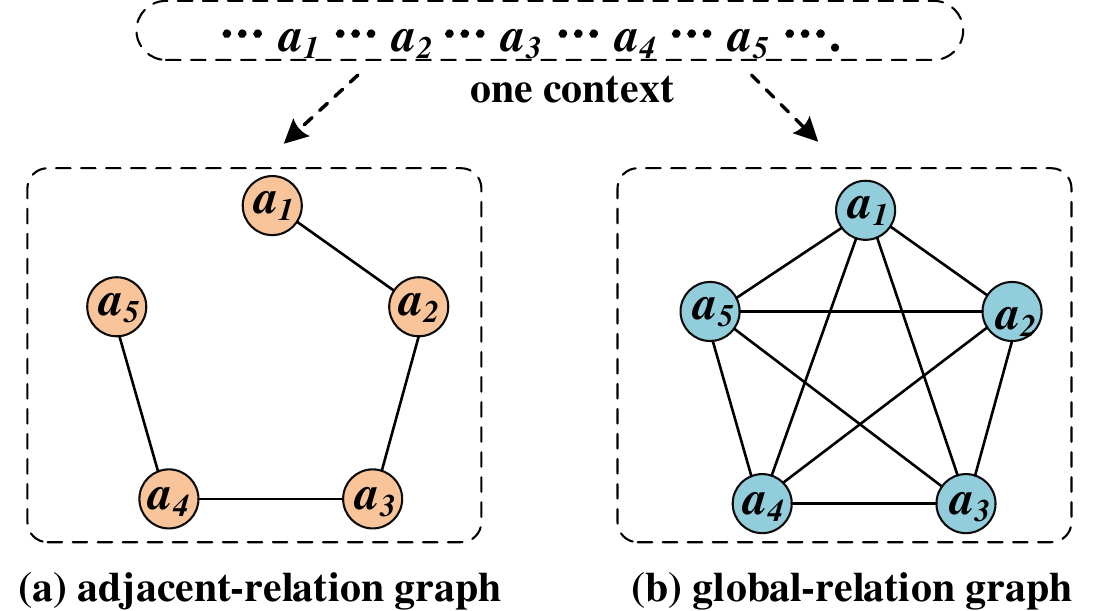}
	\caption{Illustration of our proposed sentiment graphs. $a_1$, $a_2$, $a_3$, $a_4$ and $a_5$ denote five aspects in one context.}
	\label{FIG:3}
\end{figure}

\subsubsection{Sentiment graph}
We construct a graph, named sentiment graph, to capture the sentiment dependencies between multi-aspects in one sentence, where each node is regarded as an aspect and each edge is treated as the sentiment dependency relation. As shown in Fig.~\ref{FIG:3}, we define two kinds of undirected sentiment graphs:
\begin{itemize}
\item adjacent-relation graph: An aspect is only connected to its nearby aspects.
\item global-relation graph: An aspect is connected to all other aspects.
\end{itemize}

If two nodes are connected by an edge, it means that the two nodes are neighboring to each other. Formally, given a node $v$, we use $N(v)$ to denote all neighbors of $v$.  $u\in N(v)$ means that $u$ and $v$ are connected with an edge.

\begin{figure*}
	\centering
		\includegraphics[scale=0.55]{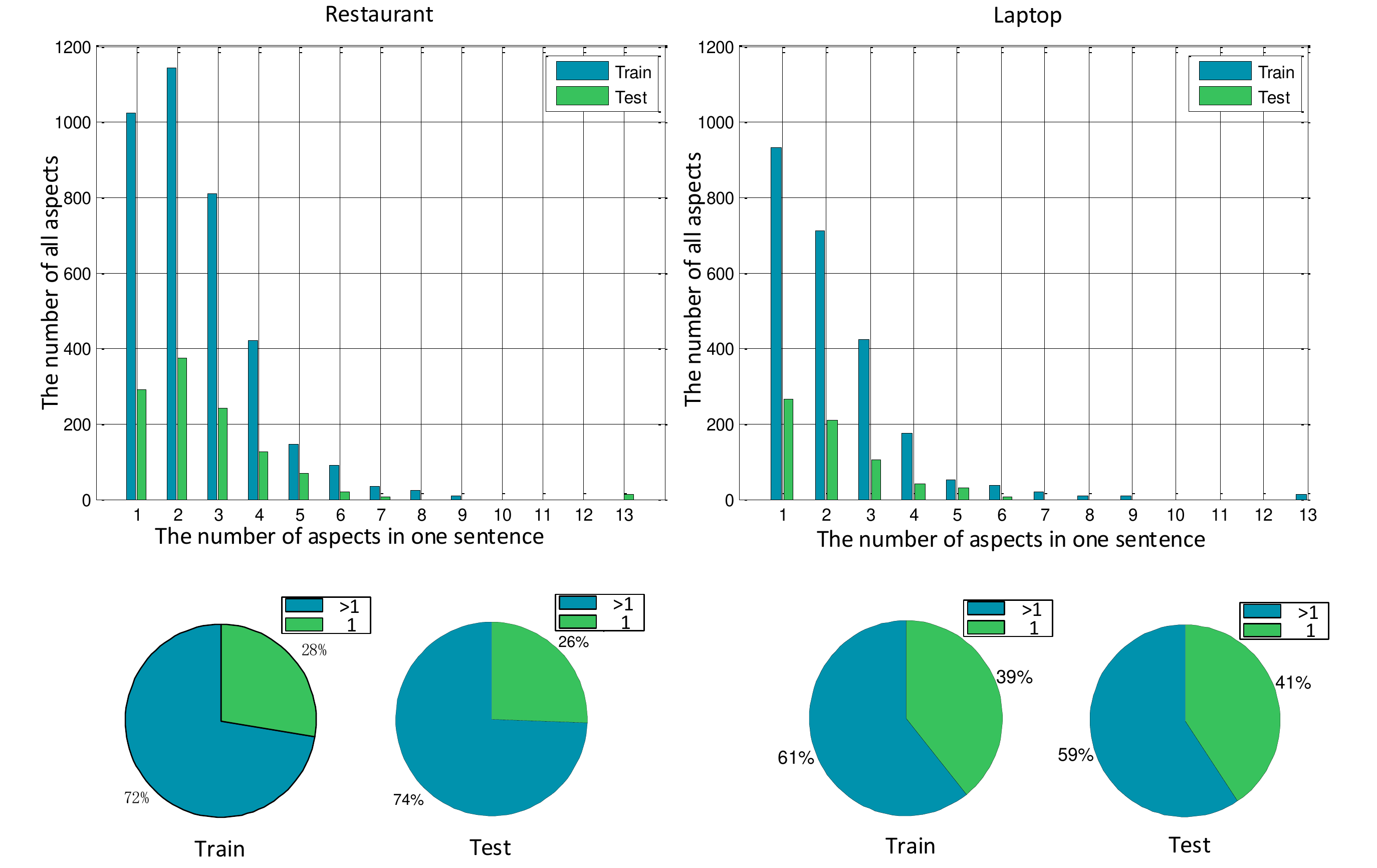}
	\caption{Statistics of the number of aspects in one sentence on SemEval 2014 data set.}
	\label{FIG:4}
\end{figure*}

\subsubsection{Sentiment graph based GCN}
GCN encodes relevant information about its neighborhood as a new representation vector, where each node in the graph indicates a representation of aspect. In addition, as Kipf et al. \cite{Kipf2016} do, we assume all nodes contain self-loops. Then, the new node representation is computed as follows:
\begin{flalign}\label{10}
&x^1_v=relu(\sum_{u\in N(v)}W_{cross}x_u+b_{cross})+ReLU(W_{self}x_v+b_{self})&
\end{flalign}
where $W_{cross}, W_{self}\in \mathbb{R}^{d_m\times d_n}$ , $b_{cross}, b_{self}\in \mathbb{R}^{d_m\times 1}$ , $x_u$ is the $u$-th aspect-specific representation (see Eq.\eqref{9}),  and $ReLU$ is the rectifier linear unit activation function. In this work, we use $d_m=d_n=2h_{hid}$.

By stacking multiple GCN layers, the final hidden representation of each node can receive messages from a further neighborhood. Each GCN layer takes the node representations from previous layer as inputs and outputs new node representations:
\begin{flalign}\label{11}
&x^{l+1}_v=relu(\sum_{u\in N(v)}W^l_{cross}x^l_u+b^l_{cross})+relu(W^l_{self}x^l_v+b^l_{self})&
\end{flalign}
where $l$ denotes the layer number and $1\leq l\leq L-1$.
\subsection{Output layer}
The final output of each GCN node $x^L_i$ is treated as a classifier of the $i$-th aspect. At last, we use a fully-connected layer to map $x^L_i$ into the aspect space of $C$ classes:
\begin{flalign}\label{12}
&z_i=W_zx^L_i+b_z&
\end{flalign}
where $W_z\in \mathbb{R}^{C\times2d_{hid}}$ is the weight matrix, and $b_z\in \mathbb{R}^{2d_{hid}\times C}$ is the bias. The predicted probability of the $i$-th aspect with sentiment polarity $j\in[1,C]$ is computed by:
\begin{flalign}\label{13}
&y'_{ij}=\frac{exp(z_{ij})}{\sum^C_{k=1}exp(z_{ik})}&
\end{flalign}

\subsection{Model training}
Our model is trained by minimizing the cross entropy with $L2$-regularization term. For a given sentence, the loss function is defined as:
\begin{flalign}\label{14}
&loss=\sum^K_{i=1}\sum^C_{j=1}y_{ij}log(y'_{ij})+\lambda\parallel\theta\parallel^2&
\end{flalign}
where $y_{ij}$ is a one-hot labels of the $i$-th aspect for the $j$-th class, $\lambda$ is the coefficient for $L2$-regularization, $\theta$ is the parameters that need to be regularized. Furthermore, we adopt the dropout strategy during training step to avoid over-fitting.

\begin{table}[tp]
    \caption{The details of the experimental data sets.}
    \label{tab:data}
    \centering
    \small
    \setlength{\tabcolsep}{5pt}
    \begin{tabular}{llccc}
    \toprule
    Data &  & Positive & Negative  & Neutral  \\
    \midrule
    \multirow{2}{*}{Restaurant}  &  Train  & 2164 & 807 & 637 \\
                                 &  Test   & 728  & 196 & 196 \\
    \cmidrule(lr){1-5}
    \multirow{2}{*}{Laptop}      &  Train  & 994  & 870 & 464 \\
                                 &  Test   & 341  & 128 & 169 \\
    \bottomrule
    \end{tabular}
\end{table}

\section{Experiments}

\begin{table*}[tp]
\caption{Comparisons with baseline models on the Restaurant dataset and Laptop dataset. The results of baseline models are retrieved from published papers. The best results in GloVe-based models and BERT-based models are all in bold separately. -A means that the model is based on adjacent-relation graph, and -G means the model is based on global-relation graph.}
\label{Tab:result}
\centering
\small
\begin{tabular}{llcccc}
\toprule
\multirow{2}{*}{Word Embedding} &\multirow{2}{*}{Models} & \multicolumn{2}{c}{Restaurant} & \multicolumn{2}{c}{Laptop}  \\ \cmidrule(lr){3-4} \cmidrule(lr){5-6}
& &  Acc   &   Macro-F1   &  Acc  &  Macro-F1      \\ \midrule

\multirow{12}{*}{GloVe}
&TD-LSTM        &75.63	&-	    &68.13	&-    \\
&ATAE-LSTM      &77.20	&-	    &68.70	&-    \\
&MenNet         &78.16	&65.83	&70.33	&64.09\\
&IAN            &78.60	&-	    &72.10	&-    \\
&RAN            &80.23	&70.80	&74.49	&\textbf{71.35}\\
&PBAN           &81.16  &-	    &74.12	&-    \\
&TSN            &80.1	&-	    &73.1	&-    \\
&AEN            &80.98	&72.14	&73.51	&69.04\\\cmidrule(lr){2-6}
&SDGCN-A w/o p  &81.61	&72.22	&73.20	&68.54\\
&SDGCN-G w/o p  &81.61	&72.93	&73.67	&68.70\\
&SDGCN-A        &82.14	&73.47	&75.39	&70.04\\
&SDGCN-G        &\textbf{82.95}	&\textbf{75.79}	&\textbf{75.55}	&\textbf{71.35}\\\midrule
\multirow{2}{*}{BERT}
&AEN-BERT	&83.12	&73.76	&79.93	&76.31\\
&SDGCN-BERT	&\textbf{83.57}	&\textbf{76.47}	&\textbf{81.35}	&\textbf{78.34}\\
\bottomrule
\end{tabular}

\end{table*}

\subsection{Data sets and experimental settings}
To demonstrate the effectiveness of our proposed method, as most previous works \cite{YequanWang2016Attention,Ma2017interactive,XiaMa2019Modeling,YouweiSong2019}, we conduct experiments on two datasets from SemEval 2014 Task4\footnote{The detailed introduction of this task can be found at \url{http://alt.qcri.org/semeval2014/task4}.} \cite{Maria2014SemEval}, which contains the reviews in laptop and restaurant. The details of the SemEval 2014 datasets are shown in Table~\ref{tab:data}. Each dataset consists of train and test set. Each review (one sentence) contains one or more aspects and their corresponding sentiment polarities, i.e., positive, neutral and negative. To be specific, the number in table means the number of aspects in each sentiment category. To demonstrate the necessity of considering the sentiment dependencies between the aspects, we further calculate the number of aspects in each sentence, which is presented in Fig.~\ref{FIG:4}. From the histogram in Fig.~\ref{FIG:4}, we can see that each sentence contains one to thirteen aspects. The number of aspects in most reviews is 1 to 4. The pie chart shows the proportion of only one aspect and more than one aspect in one sentence. It can be seen that more than half of the aspects do not appear alone in a review. According to these statistics, we can conclude that it is common to have multi-aspects within one sentence. Our model mainly aims to model the sentiment dependencies between different aspects in one sentence.

In our implementation, we respectively use the GloVe\footnote{https://nlp.stanford.edu/projects/glove/} \cite{Pennington2014Glove} word vector and the pre-trained language model word representation BERT\footnote{https://github.com/google-research/bert\#pre-trained-models } \cite{Devlin2019BERT} to initialize the word embeddings.  The dimension of each word vector is 300 for GloVe and 768 for BERT. The number of LSTM hidden units is set to 300, and the output dimension of GCN layer is set to 600. The weight matrix of last fully connect layer is randomly initialized by a normal distribution $N(0, 1)$. Besides the last fully connect layer, all the weight matrices are randomly initialized by a uniform distribution $U(-0.01, 0.01)$. In addition, we add $L2$-regularization to the last fully connect layer with a weight of 0.01. During training, we set dropout to 0.5, the batch size is set to 32 and the optimizer is Adam Optimizer with a learning rate of 0.001. We implement our proposed model using Tensorflow\footnote{https://www.tensorflow.org/}. To evaluate performance of the model, we employ Accuracy and Macro-F1 metrics. The Macro-F1 metric is more appropriate when the data set is not balanced.
\subsection{Comparative methods}

To comprehensively evaluate the performance of proposed SDGAN, we compare our model with the following models.
\begin{itemize}
\item \textbf{TD-LSTM} \cite{DuyuTang2016} constructs aspect-specific representation by the left context with aspect and the right context with aspect, then employs two LSTMs to model them respectively. The last hidden states of the two LSTMs are finally concatenated for predicting the sentiment polarity of the aspect.
\item \textbf{ATAE-LSTM} \cite{YequanWang2016Attention} first attaches the aspect embedding to each word embedding to capture aspect-dependent information, and then employs attention mechanism to get the sentence representation for final classification.
\item \textbf{MemNet} \cite{DuyuTang2016Aspect} uses a deep memory network on the context word embeddings for sentence representation to capture the relevance between each context word and the aspect. Finally, the output of the last attention layer is used to infer the polarity of the aspect.
\item \textbf{IAN} \cite{Ma2017interactive} generates the representations for aspect terms and contexts with two attention-based LSTM network separately. Then the context representation and the aspect representation are concatenated for predicting the sentiment polarity of the aspect.
\item \textbf{RAM} \cite{chenPeng2017} employs a gated recurrent unit network to model a multiple attention mechanism, and captures the relevance between each context word and the aspect. Then the output of the gated recurrent unit network is obtained for final classification.
\item \textbf{PBAN} \cite{Gu2018A} appends the position embedding into each word embedding. It then introduces a position-aware bidirectional attention network (PBAN) based on Bi-GRU to enhance the mutual relation between the aspect term and its corresponding sentence.
\item \textbf{TSN} \cite{XiaMa2019Modeling} is a two-stage framework for aspect-level sentiment analysis. The first stage, it uses a position attention to capture the aspect-dependent representation. The second stage, it introduces penalization term to enhance the difference of the attention weights towards different aspects in one sentence.
\item \textbf{AEN} \cite{YouweiSong2019} mainly consists of an embedding layer, an attentional encoder layer, an aspect-specific attention layer, and an output layer. In order to eschew the recurrence, it employs attention-based encoders for the modeling between the aspect and its corresponding context.
\item \textbf{AEN-BERT} \cite{YouweiSong2019} is AEN with BERT embedding.
\end{itemize}

\subsection{Overall results}
Table~\ref{Tab:result} shows the experimental results of competing models. In order to remove the influence with different word representations and directly compare the performance of different models, we compare GloVe-based models and BERT-based models separately. Our proposed model achieves the best performance on both GloVe-based models and BERT-based models, which demonstrates the effectiveness of our proposed model. In particularly, SDGCN-BERT obtains new state-of-the-art results.

Among all the GloVe-based methods, the TD-LSTM approach performs worst because it takes the aspect information into consideration in a very coarse way. ATAE-LSTM, MenNet and IAN are basic attention-based models. After taking the importance of the aspect into account with attention mechanism, they achieve a stable improvement comparing to the TD-LSTM. RAM achieves a better performance than other basic attention-based models, because it combines multiple attentions with a recurrent neural network to capture aspect-specific representations. PBAN achieves a similar performance as RAM by employing a position embedding. To be specific, PBAN is better than RAM on Restaurant dataset, but worse than RAN on Laptop dataset. Compared with RAM and PBAN, the overall performance of TSN is not perform well on both Restaurant dataset and Laptop dataset, which might because the framework of TSN is too simple to model the representations of context and aspect effectively. AEN is slightly better than TSN, but still worse than RAM and PBAN. It indicates that the discard of the recurrent neural networks can reduce the size of model while lead to the loss of performance.

Comparing the results of SDGCN-A w/o position and SDGCN-G w/o position, SDGCN-A and SDGCN-G, respectively, we observe that the GCN built with global-relation is slightly higher than built with adjacent-relation in both accuracy and Macro-F1 measure. This may indicate that the adjacent relationship is not sufficient to capture the interactive information among multiple aspects due to the neglect of the long-distance relation of aspects. Moreover, the two models (SDGCN-A and SDGCN-G) with position information gain a significant improvement compared to the two models without position information. It shows that the position encoding module is crucial for good performance.

Benefits from the power of pre-trained BERT, BERT-based models have shown huge superiority over GloVe-based models. Furthermore, compared with AEN-BERT, on the Restaurant dataset, SDGCN-BERT achieves absolute increases of 1.09\% and 1.86\% in accuracy and Macro-F1 measure respectively, and gains absolute increases of 1.42\% and 2.03\% in accuracy and Macro-F1 measure respectively on the Laptop dataset. The increments prove the effectiveness of our proposed SDGCN.

\begin{figure}
	\centering
		\includegraphics[scale=.52]{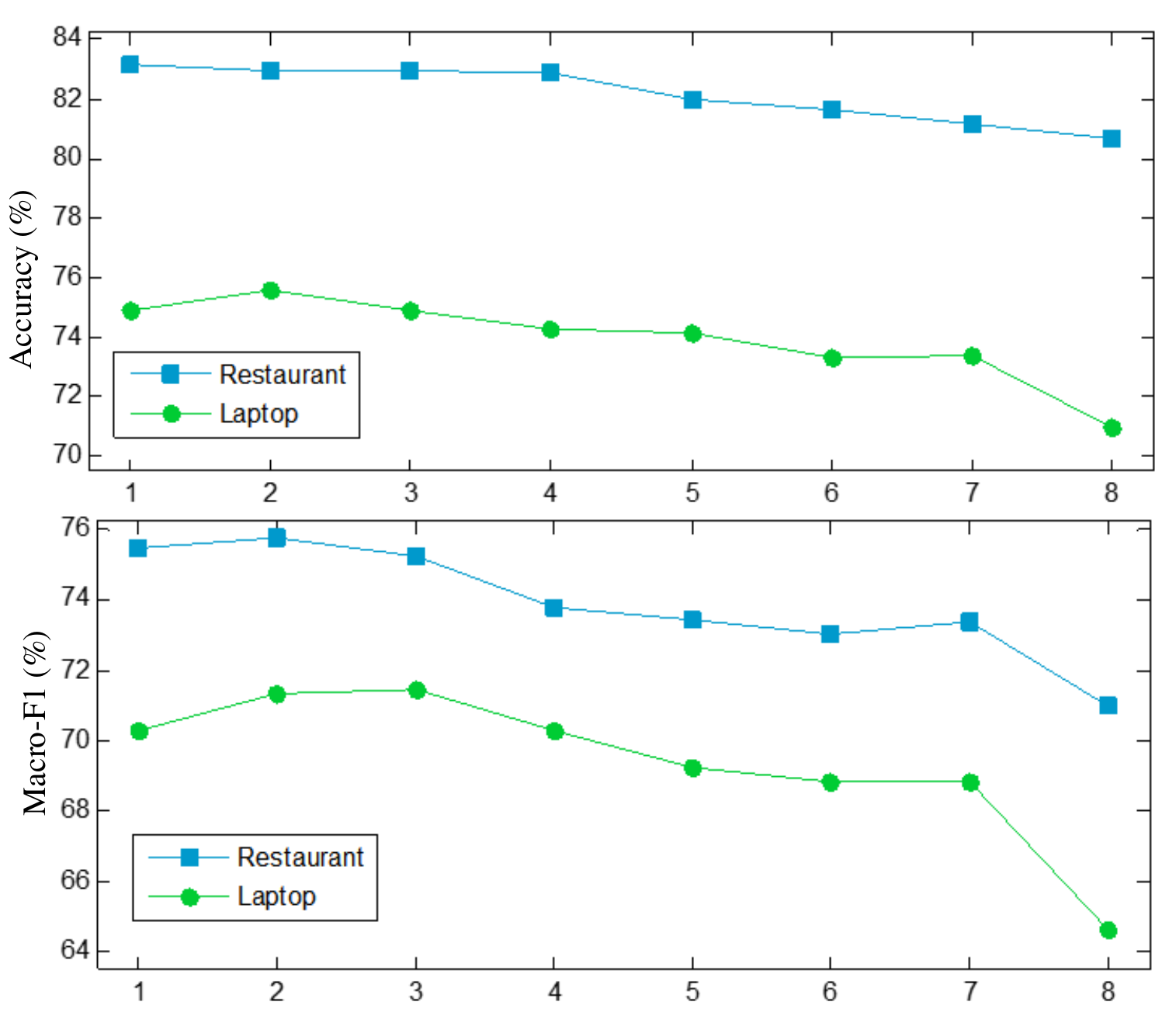}
	\caption{Comparisons with different depths of GCN in our model.}
	\label{FIG:5}
\end{figure}

\begin{figure*}
	\centering
		\includegraphics[scale=1.05]{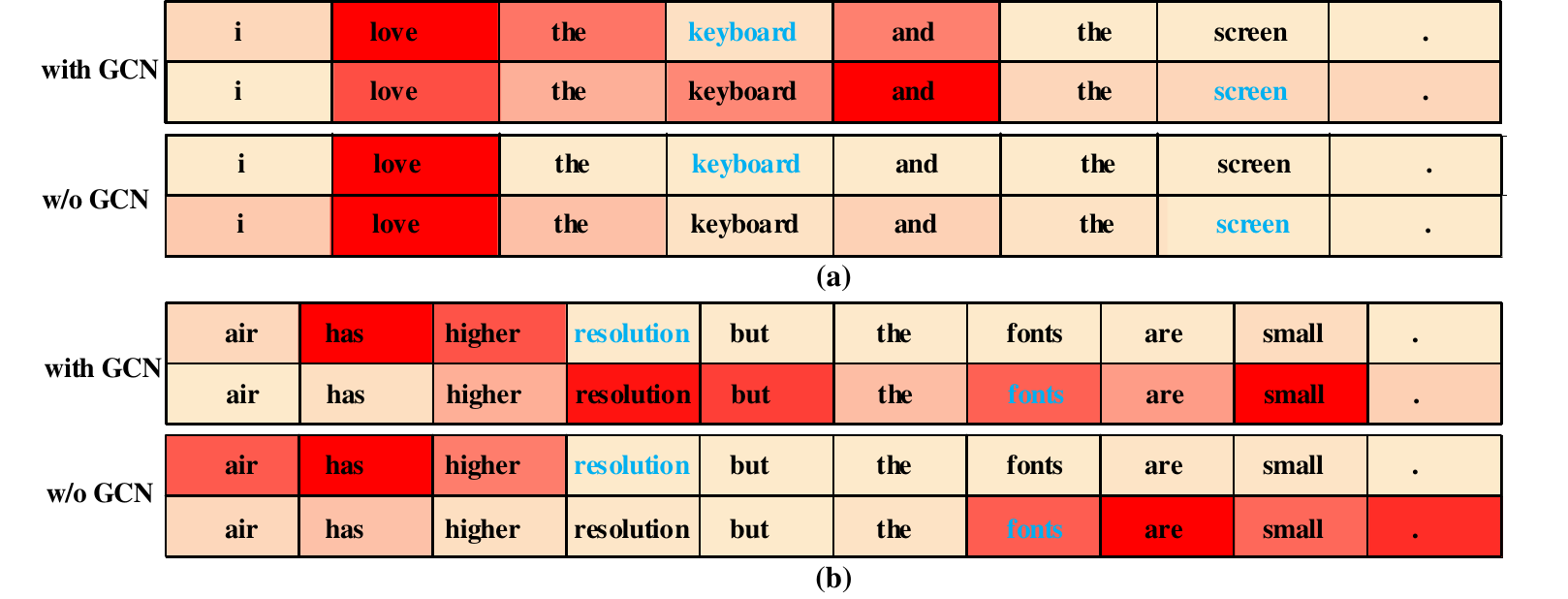}
	\caption{Illustration of attention weights obtained by model with GCN and without GCN respectively. (a) and (b) are two examples from the Laptop dataset. (a) Aspects: keyboard, screen; (b) Aspects: resolution, fonts.}
	\label{FIG:6}
\end{figure*}

\begin{table}[hp]
\caption{The effect of GCN.}
\label{Tab:GCN}
\begin{adjustbox}{center}
\small
\setlength{\tabcolsep}{4pt}
\begin{tabular}{lcccc}
\toprule
\multirow{2}{*}{Models}   & \multicolumn{2}{c}{Restaurant}  & \multicolumn{2}{c}{Laptop} \\ \cmidrule(lr){2-3} \cmidrule(lr){4-5}
                                         & Acc & Macro-F1 & Acc & Macro-F1   \\
\midrule
\multicolumn{1}{l}{Att}                 & 81.43 & 72.40 & 72.12 & 68.67  \\ 
\multicolumn{1}{l}{Att+GCN}             & 82.77 & 74.33 & 74.61 & 70.33  \\ \cmidrule(lr){1-5}
\multicolumn{1}{l}{BiAtt}               & 81.61 & 73.49 & 73.51 & 69.73 \\ 
\multicolumn{1}{l}{BiAtt+GCN (SDGCN)}   & 82.95 & 75.79 & 75.55 & 71.35 \\ 

\bottomrule
\end{tabular}
\end{adjustbox}
\end{table}

\subsection{The effect of GCN module}

In this section, we design a series of models to further verify the effectiveness of GCN module. These models are:
\begin{itemize}
\item \textbf{BiAtt+GCN} is just another name of our proposed SDGCN model.
\item \textbf{BiAtt} is based on BiAtt -GCN, where we remove the GCN module. Therefore, it predicts the sentiments of different aspects in one sentence independently.
\item \textbf{Att+GCN} is a simplified version of BiAtt+GCN. The only difference between Att+GCN and BiAtt+GCN is that Att+GCN does not have \textit{context to aspect attention}.
\item \textbf{Att} is the model of Att+GCN removing the GCN module.
\end{itemize}

Table~\ref{Tab:GCN} shows the performances of all these models. It is clear to see that, comparing with GCN-reduced models, the two models with GCN achieve higher performance, respectively. The results verify that the modeling of the sentiment dependencies between different aspects with GCN plays a great role in predicting the sentiment polarities of aspects.

\subsection{Impact of GCN layer number}
The number of GCN layers is one very important setting parameter that affects the performance of our model. In order to investigate the impact of the GCN layer number, we conduct experiment with the different number of GCN layers from 1 to 8. The performance results are shown in Fig.~\ref{FIG:5}. As can be seen from the results, in general, when the number of GCN layers is 2, the model works best. When the number of GCN layers is bigger than 2, the performance drops with the increase of the number of GCN layers on both the datasets. The possible reason for the phenomenon of the performance drop may be that with the increase of the model parameters, the model becomes more difficult to train and over-fitting.

\subsection{Case study}
In order to have an intuitive understanding of the difference between with-GCN model (our proposed model) and without-GCN model, we use two examples with multiple aspects from laptop dataset as a case study. We draw heat maps to visualize the attention weights on the words computed by the two models, as shown in Fig.~\ref{FIG:6}. The deeper the color, the more attention the model pays to it.

As we can see from the first example, i.e., \textit{``i love the keyboard and the screen."}, with two aspects \textit{``keyboard"} and \textit{``screen"}, without-GCN model mainly focuses on the word \textit{``love"} to predict the sentiment polarities of the two aspects. While for with-GCN model, besides the word \textit{``love"}, it also pays attention to the conjunction \textit{``and"}. This phenomenon indicates that with-GCN model captures the sentiment dependencies of the two aspects through the word \textit{``and"}, and then predicts the sentiments of \textit{``keyboard"} and \textit{``screen"} simultaneously.

The second example is \textit{``air has higher resolution but the fonts are small."} with two aspects \textit{``resolution"} and \textit{``fonts"}. It is obvious that the sentiments of the two aspects \textit{``resolution"} and \textit{``fonts"} are opposite connected by the conjunction \textit{``but"}. Without-GCN model predicts the polarity of aspect \textit{``resolution"} by the word \textit{``higher} and the polarity of aspect \textit{``fonts} by the word \textit{``small"} in isolation, which ignores the relation between the two aspects. In the contrary, with-GCN model enforces the model to pay attention on the word \textit{``but"} when predicting the sentiment polarity for aspect \textit{``fonts"}.

From these examples, we can observe that our proposed model (with-GCN model) not only focuses the corresponding words which are useful for predicting the sentiment of each aspect, but also considers the textual information which is helpful for judging the relation between different aspects. By using attention mechanism to focus on the textual words describing the interdependence between different aspects, the downstream GCN module can effectively further represent the sentiment dependencies between different aspects in one sentence. With more useful information, our proposed model can predict aspect-level sentiment category more accurately.	
\section{Conclusion}
In this paper, we design a novel GCN based model (SDGCN) for aspect-level sentiment classification. The key idea of our model is to employ GCN to model the sentiment dependencies between different aspects in one sentence. Specifically, SDGCN first adopts bidirectional attention mechanism with position encoding to obtain aspect-specific representations, then captures the sentiment dependencies via message passing between aspects. Thus, SDGCN benefits from such dependencies which are always ignored in previous studies. Experiments on SemEval 2014 verify the effectiveness of the proposed mode, and SDGCN-BERT obtains new state-of-the-art results. The case study shows that SDGCN can not only pay attention to those words which are important for predicting the sentiment polarities of aspects, but also pay attention to the words which are helpful for judging the sentiment dependencies between different aspects.

In our future work, we will explore how to build a more precise sentiment graph structure between aspects. The two kinds of undirected sentiment graphs in this work are coarse. We conjecture that making use of textual information to define a graph may create a better graph structure.



\biboptions{numbers,sort&compress}
\bibliographystyle{elsarticle-num}
\bibliography{sdgcn}

%
%
%
%
\end{document}